\documentclass{article}
\PassOptionsToPackage{numbers, compress}{natbib}
\usepackage[preprint]{arXiv}
\usepackage[utf8]{inputenc} 
\usepackage[T1]{fontenc}    
\usepackage{hyperref}       
\usepackage{url}            
\usepackage{booktabs}       
\usepackage{amsfonts}       
\usepackage{nicefrac}       
\usepackage{microtype}      
\usepackage{graphicx}
\usepackage{amsmath}
\usepackage{multirow}
\usepackage{array}
\usepackage[noend]{algpseudocode}
\usepackage{algorithmicx,algorithm}
\graphicspath{ {images/} }

\title{FashionVQA: \\A Domain-Specific Visual Question Answering System}

\author{
  Min Wang \\
  Target Corporation\\
  100 S. Mathilda Place Suite 300 \\
  Sunnyvale, CA 94086\\
  \texttt{min.wang@target.com} \\
  \And
  Ata Mahjoubfar \\
  Target Corporation\\
  100 S. Mathilda Place Suite 300 \\
  Sunnyvale, CA 94086\\
   \texttt{ata.mahjoubfar@target.com}
  \AND
  Anupama Joshi \\
  Target Corporation\\
  100 S. Mathilda Place Suite 300 \\
  Sunnyvale, CA 94086\\
   \texttt{anupama.joshi@target.com}
 }

\begin{document}

\maketitle

\begin{abstract}
Humans apprehend the world through various sensory modalities, yet language is their predominant communication channel. Machine learning systems need to draw on the same multimodal richness to have informed discourses with humans in natural language; this is particularly true for systems specialized in visually-dense information, such as dialogue, recommendation, and search engines for clothing. To this end, we train a visual question answering (VQA) system to answer complex natural language questions about apparel in fashion photoshoot images. The key to the successful training of our VQA model is the automatic creation of a visual question-answering dataset with 168 million samples from item attributes of 207 thousand images using diverse templates. The sample generation employs a strategy that considers the difficulty of the question-answer pairs to emphasize challenging concepts. Contrary to the recent trends in using several datasets for pretraining the visual question answering models, we focused on keeping the dataset fixed while training various models from scratch to isolate the improvements from model architecture changes. We see that using the same transformer for encoding the question and decoding the answer, as in language models, achieves maximum accuracy, showing that visual language models (VLMs) make the best visual question answering systems for our dataset. The accuracy of the best model surpasses the human expert level, even when answering human-generated questions that are not confined to the template formats. Our approach for generating a large-scale multimodal domain-specific dataset provides a path for training specialized models capable of communicating in natural language. The training of such domain-expert models, e.g., our fashion VLM model, cannot rely solely on the large-scale general-purpose datasets collected from the web.
\end{abstract}

\section{Introduction}
Fashion is about 2\% of the world's GDP and a significant sector of the retail industry. Whenever a new fashion item like apparel or footwear is launched, the retailer needs to prepare and show rich information about the product, including pictures, text descriptions, and detailed attribute tags. The attributes of the fashion products, including color, pattern, texture, material, occasion-to-use, etc., require domain experts to label them piece by piece. This labeling process is time-consuming, costly, subjective, error-prone, and fundamentally imprecise due to the interdependency of the attributes. To address these issues, we introduce a multi-task multimodal machine learning model to automatically, consistently, and precisely infer the visual attributes of the fashion items.

Each item is typically labeled with multiple tags that describe different attributes of the item. For example, an item can be labeled with ``shirt'', ``red'', ``solid pattern'', ``blue collar'' and ``short sleeve''. An intuitive way of learning such information is to train a multi-label classifier, which outputs the probability of multiple labels of each input sample. However, such a model cannot encode the relationship between different attributes. For example, ``short sleeve'' is a suitable attribute for ``shirt'', but not for ``jeans'', and ``red'' only describes the body part of the shirt, but not the collar. The model needs to learn attribute and object relationships and adjusts its output accordingly.

We propose designing a Visual Questioning Answer (VQA) framework for fashion items, in which the model is trained to answer complex natural-language questions, such as ``is the person wearing a red shirt with solid pattern and blue collar?'', given the input image. The VQA task is more challenging than the simple attribute classifier since it requires a thorough understanding of both the question and the structure and relationship between various visual attributes in the image. By training such a model, we convert the manual process of tagging new products with visual attributes into automated answering of a series of questions with visual intents (auto-labeling). The model also generates multimodal embeddings of the product images attended to the questions for downstream dialogue, search, and recommendation systems.

Prior to our work, there exists a large-scale VQA v2 dataset \cite{vqa2}, which includes 0.6 million \textit{question-answer-image} triplets. It has been widely used as the benchmark in recent research on VQA tasks. However, this general dataset only contains a small number of \textit{question-answer-image} triplets related to fashion. In this work, we build a fashion VQA dataset from a diverse apparel product database. The questions, including both binary and non-binary, are automatically composed by filling question templates with the given attribute information. The dataset contains 207 thousand images and 168 million \textit{question-answer-image} triplets. The automatic generation of the VQA dataset from a limited number of images and attributes allows us to achieve the scale required for training a multimodal domain expert model.

We leverage a cross-modality fusion model mapping representations from visual and text space to the same latent feature space and performing answer prediction with classifier modules. Given an image that contains a fashion item and the corresponding questions regarding its different attributes, the model predicts the answers to the given questions. We can then use the model to generate the missing or alternative attribute information based on its answers.

Additionally, given different but similar text descriptions on the same item, we can generate consistent feature embeddings that enable us to build better online search services. The existing search engines cannot attend to the relevant visual parts of a fashion item given the query and do not adapt the attention mask according to the chained adjectives. With this work, we can map the input query to the learned embeddings space and perform a robust and fuzzy search in that multimodal space. We can also provide a visual dialogue service, in which the customers can ask consecutive questions to narrow down the item list according to their apparel preferences. We can also build a fashion recommendation system in the multimodal embeddings space. The customer-item interaction history is mapped to this space, and the neighboring items are recommended.

\section{Related work}
\textbf{Visual feature learning in VQA:}
The visual feature vector is often extracted from the input image using a Convolutional Neural Networks (CNN) as the visual encoder, e.g., VGG \cite{vgg}, ResNet \cite{resnet}, or ResNext \cite{resnext} models. In the early VQA frameworks, grid-based visual features extracted by ImageNet-pretrained \cite{imagenet} VGG or ResNet models were widely adopted. Since \cite{bottom-up-attention}, region-based visual features extracted using Faster R-CNN \cite{faster-rcnn}-based object detection model, especially fine-tuned on Visual Genome \cite{visualgenome} dataset, have been dominant \cite{dfaf}\cite{mcan}\cite{oscar}\cite{lxmert}. In \cite{grid-feature}, the authors propose extracting the grid features from the same layer as the pretrained detector, achieving comparable performance as the region-based features with higher efficiency. We benchmark these two types of visual feature extraction methods on our dataset across different VQA models.

\textbf{Cross-modality fusion models:}
Cross-modality fusion model is a core component of the VQA framework. It aligns the features from the visual and language modalities. Initially-proposed VQA models identify the high-level cross-modal interactions by Bilinear Fusion \cite{bilinear}. MCB \cite{mcb}, MLB \cite{mlb} and MUTAN \cite{mutan} are later introduced to achieve better fusion performance at much lower computational cost and parameters. Motivated by the remarkable performance of the attention mechanism in language and vision models \cite{bert}\cite{attention}, the attention module becomes the fundamental block in designing the cross-modality fusion models. DFAF \cite{dfaf} uses self-attention and co-attention modules to learn the inter-and intra-connections between the two modalities. MCAN \cite{mcan} builds the model with the blocks of self-attention and guided attention. LXMERT \cite{lxmert}, VilBERT \cite{vilbert} adopt a similar strategy and build a two-stream co-attention-based model. VisualBERT \cite{visualbert}, UNITER \cite{uniter}, OSCAR \cite{oscar}, OSCAR+ \cite{oscar+} learn the alignment between image and language by pretraining on multiple image caption datasets with BERT-style \cite{bert} visual language models (VLMs).

\textbf{Fashion datasets:} 
In recent year, many valuable fashion datasets \cite{deepfashion}\cite{deepfashion2}\cite{modanet}\cite{fashionai}\cite{fashioniq}\cite{imaterialist} \cite{fashionpedia}\cite{fashion1}\cite{fashion2}\cite{runway}\cite{fashion-mnist} have greatly contributed to clothing item recognition and apparel attribute understanding. However, most of them suffer from some limitations when considered for training versatile VQA models. In \cite{runway}\cite{modanet}\cite{fashion-mnist}\cite{fashion2}, only primary categories in the dataset are labeled. Additional garment parts and attributes are annotated in \cite{imaterialist}\cite{deepfashion2}\cite{fashionai}. Segmentation masks over each piece of garment are drawn for the semantic segmentation task in \cite{modanet}\cite{runway}\cite{deepfashion2}\cite{fashionpedia}. Generally, the localization of the garment pieces and parts in these types of datasets takes considerable human annotation labor, and few of the datasets are suitable for conversion to a new dataset for vision language tasks.

\section{Methods}
In this section, we describe how we designed and generated a novel VQA dataset for fashion. We named the new dataset FashionVQA dataset.   
\subsection{Terminologies}
\textbf{Category:} Each clothing item can be labeled with one super-category and several primary categories or sub-categories. 
\begin{itemize}
  \item Super-categories: ``apparel top'', ``apparel bottom'', ``one-piece clothing'', ``shoes'', and ``accessories''.
  \item Primary categories: ``shirt'', ``sweater'', ``jacket'', ``pants'', ``skirt'', ``dress'', ``jumpsuit'', ``boots'', ``sneaker'', ``gloves'', etc.
  \item Sub-categories: ``t-shirt'', ``cardigan'', ``blazer jacket'', ``pencil skirt'', ``sweatpants'', ``overall jumpsuit'', ``hiking boots'', etc.
\end{itemize}

\textbf{Attributes:} ``Color'', ``pattern'', ``fit type'', ``closure type'', and ``material/fabric'' are the general attributes for all the fashion items. Each apparel type also has its unique attributes. The unique attributes for ``apparel top'', ``apparel bottom'', ``one-piece clothing'', and ``shoes'' are listed below.
\begin{itemize}
  \item Apparel top: ``torso length type'', ``sleeve length type'', ``pocket type'', ``neckline type'', ``sleeve style'', ``collar type'', ``lapel type'', and ``cuff type''.
  \item Apparel bottom: ``pant leg type'', ``skirt length type'', ``pant leg style'', and ``pleat type''.
  \item One-piece clothing: ``neckline type'', ``sleeve length type'', ``sleeve style'', ``pant leg type'', ``skirt length type'', and ``pleat type''. 
  \item Shoes: ``height'', ``width'', ``toe openness'', and ``shape of toe''.
\end{itemize}

\textbf{Attribute values:} Each attribute is composed of a set of \textit{attribute values}. For example, the set of the \textit{attribute values} of color attribute includes ``red'', ``black'', ``green'',``blue'', ``yellow'', etc.

\textbf{Parts:} The \textit{parts} mentioned in our dataset are typically lined on the fashion item, such as ``patches'' and ``pockets''.

\textbf{Location:} In our dataset, there exist numerous images with a person wearing multiple fashion items. Therefore, we use \textit{Location} to specify the relative location of the primary fashion item in the image, such as ``on the top'', ``on the bottom'', ``on the feet'', ``over the neck'', or ``on the head''.

\subsection{Data Collection pipeline}
Our data collection pipeline involves four steps: [1] querying fashion items' unique identity numbers (image IDs), [2] querying and parsing meta-information, [3] downloading images, and [4] filling question templates and forming \textit{question-answer-image} triplets.

Each fashion item comes with a unique identity (ID) number. First, we query all fashion items and retrieve their IDs. Then, we predefine a data structure that is eligible to query the meta-information of fashion items from the item database. Feeding the data structure to an open-source data query API, ``graphQL'', we can obtain the meta-information attached to each ID, which contains the primary image (front-view) URL and the description of the primary fashion item. We can directly download the primary image from the URL with Python.

The description of the fashion item is not an on-deck dictionary that maps each unique targeted attribute to its corresponding set of \textit{attribute values}. For example, ``Color'' could be described with different phrases such as ``Product Color'' or ``Color Name''. Parsing from the description is a process that collects \textit{attribute values} from various sources and reduces similar attribute terminologies into the same group. Also, meta-information comes in a very raw manner with many \textit{attribute values} cross \textit{attributes} entangled, e.g., ``black/stripes'', or in a vague expression. In this stage, we also need to clean these \textit{attribute values} and map them into common terminologies, e.g., map ``black/stripes'' to ``black'' for color and ``stripes'' for pattern, or ``olive night'' color to ``olive green''. 

\subsection{Question templates}
We adopt a templating mechanism to automatically create \textit{question-answer} pairs. The question templates are designed based on a set of fixed rules that meet the English grammar and result in human-readable sentences. By filling the question templates with specific item \textit{attribute}, \textit{attribute value}, \textit{category}, and \textit{location}, we can generate a variety of questions for each image. Answer of each question can be \textit{``Yes/No''} for binary questions and multiple choices from the relevant \textit{attribute values} for non-binary questions.

Since the images from the FashionVQA dataset are all photoshoot images with a solid background, the question templates ask only attribute-related questions about the fashion items in the image. For example, ``what is the sleeve length of this shirt on the top?'' or ``is this a white v-neck sweater?''. The basic template is structured as ``\{\textit{question type}\} \{this/these\} \{a/an/\} \{pair of/pairs of/\} \{\textit{object}\} \{\textit{location}\}?''. When filling the template to expand into a full sentence, the choices between ``is/are'', ``this/these'', ``a/an'', ``a pair of/pairs of'', and singular or plural format of \textit{category} are required to follow the English grammar and be aligned with the number of targeted fashion item in the image. For example, if the \textit{number of pieces} in the image is more than one, we choose ``are'', ``these'', `` /pairs of'', and plural format of \textit{category}. If the fashion item includes pant legs or two pieces like eyeglasses, we add ``pair of / pairs of'' in the question templates. 

If a person is in an image, and the primary fashion item is not from the super-category of ``one-piece clothing'', we assume there are multiple fashion items in the image. We use ``\{\textit{location}\}'' to specify the relative location of the primary fashion item. We use ``on the top'' for ``apparel top'', ``on the bottom'' for ``apparel bottom'', ``on the feet'' for ``shoes'', ``on the head'' for ``hat'', and ``over the neck'' for ``scarf''.

The question templates fall into two primary categories based on the answer types: binary and non-binary templates. 

\textbf{Binary question templates:} Binary question templates typically start with ``is this/are these'', ``can you see'', or ``is there any \{\textit{part}\} on this/these'', followed by the description of the targeted item in the format of ``\{\textit{location}\} \{a\}/\{a pair of/\}/\{\} \{\textit{attribute value 1}\} \{\textit{attribute value 2}\} \{\textit{category}\} '', whereas \textit{attribute value 1} and \textit{attribute value 2} are two \textit{attribute values} from different \textit{attributes}. Permuting \textit{attribute value 1}, \textit{attribute value 2}, \textit{category} in different orders yields different question templates. Conjunction words like ``with'', ``and'', or ``in'' can be used in templates when \textit{attribute value 1} or \textit{attribute value 2}, or both are located after \textit{category}. The most common question types used in binary questions are ``is/are'' and ``can''.  

\textbf{Non-binary question templates:} Non-binary question templates typically start with question words like ``what'' / ``why'' / ``when'' / ``how'' followed by terms of attribute. The formats of the question type vary from attribute to attribute. For example, the question type can be ``what color is'' or ``what is the color of'' for attribute ``color'', ``what pattern is on'' or ``what print is on'' for attribute ``pattern'', and ``how many'' or ``what number of'' for attribute ``number of pockets''.

Unlike binary question templates in our current dataset, we do not leverage other \textit{attribute values} unrelated to the targeted attribute in filling the non-binary question templates; even the \textit{category} of the targeted fashion item is not necessary. Therefore, it is possible to increase the diversity of the non-binary question templates with additional \textit{attribute values} or \textit{categories}. For example, we can come up with a color question template like ``what color is on the top?'' or ``what color is this shirt the person wearing on the top?''.

\textbf{Diversification:} The primary question templates are those preserving all the \textit{demonstratives}, \textit{subject pronouns}, and \textit{prepositional phrases}. By randomly either removing parts of those phrases or replacing them with alternatives, we can create assorted variant question templates.

In non-binary question templates, the question types for a given attribute come in different fashions, contributing to diverse non-binary question templates. Additionally, it is reasonable to replace the specific \textit{category} information of the targeted item with the combination of \textit{pronoun} and \textit{location} to expand the diversity of question templates. Adding non-relevant \textit{attribute values} to describing the fashion item is also an approach to creating new question sentences. To further increase the robustness of the question templates, we also introduce a small portion of noise into the question templates, switching between ``this/these'', ``is/are'', and ``singular/plural''.

In binary question templates, even elimination of the question type phrases like ``is this a'', ``are these'', or ``is there'' does not cause an obstacle to make the remaining phrase human readable. Therefore, we truncate a small fraction of the full question sentences by removing phrases of question type to increase the diversity of the binary questions. When \textit{attribute values} are placed after the \textit{category}, we randomly pick one from different  \textit{conjunction structures} to form different phrases, which will remarkably increase the diversity of the binary question sentences. For example, for ``a shirt with stripe pattern'', an alternative expression can be ``a shirt designed with stripe pattern'', or ``a shirt featured in stripe design''.

Table~\ref{qt_example} demonstrates some examples of question sentences generated from question templates. 
\begin{table}
  \centering
  \scalebox{0.86}{
  \begin{tabular}{|p{0.42\linewidth}|p{0.11\linewidth}|p{0.12\linewidth}|p{0.38\linewidth}|} \hline 
    Question templates & Answer types & Question types & Questions \\ \hline 
    ``is this a \{\textit{attr1}\} \{\textit{category}\} with \{\textit{attr2}\}?'' & ``yes/no'' & ``is/are'' & ``is this a white shirt with long sleeves?'' \\ \hline
    ``on the top a \{\textit{category}\} with \{\textit{attr1}\} and in \{\textit{attr2}\} design?'' & ``yes/no'' & ``is/are'' & ``on the top a sweater with floral print and in v neck design?'' \\ \hline 
    ``what \{\textit{attribute}\} is this \{\textit{category}\} the person wearing \{\textit{location}\}?''
&``others'' &
``what \newline\{\textit{attribute}\}'' &
``what color is this a-line dress the person wearing on the top?'' \\ \hline
   ``what \{\textit{attribute}\} is the one \{\textit{location}\}?''
&``others'' &
``what \newline\{\textit{attribute}\}'' &
``what color is the one on the top?'' \\ \hline 
    ``when is a good time to wear this \{\textit{attr1}\} \{\textit{category}\}?''
&``others'' &
``when'' &
``when is a good time to wear this yellow dress?'' \\ \hline
  \end{tabular}}
  \caption{Question templates and examples in the FashionVQA dataset}
  \label{qt_example}
\end{table}

\subsubsection{Balance positive and negative samples for each binary question}
Given binary and non-binary question templates and \textit{attribute values} for a specific image, we can easily generate non-binary \textit{question-(multiple answers)-image} triplets and binary \textit{question-(positive answer)-image} triplets. 

For a balanced VQA dataset, we expect each binary question to come with the same number of positive and negative samples, i.e., balanced \textit{(question, ``Yes'', image ID)} triplets and \textit{(question, ``No'', image ID)} triplets. Here, we consider two different strategies for generating the negative samples of each binary question. One strategy keeps the image fixed and changes the \textit{attribute values} in the question; the other one keeps the \textit{attribute values} fixed and changes the image. Here we further explain these strategies in detail:

\textbf{Image-based:} For each image, by filling the binary question templates with specific \textit{attribute values} and \textit{category} information provided for this image, we make a positive binary sample. When an \textit{attribute value} or \textit{category} in an existing binary question is changed, if the alteration is not in the list of \textit{attribute values} or \textit{categories} corresponding to the image, we assume this is a negative sample for the binary question.
  
\begin{algorithm}[h]
\caption{Attribute-based balancing of the positive and negative samples for binary questions}
\label{alg:balanced}
\hspace*{0.02in} {\bf Input:} $\operatorname{S}$: \{$s_i, \ldots$\} list of all fashion items \\
\hspace*{0.02in} Each fashion item $s_i$: \{image ID: $u_i$, category: $c_i$, attributes: \{$a_k, \ldots$\}, attribute values: \{$v_{a_k}, \ldots$\} d\} \\
\hspace*{0.02in} $\operatorname{Q_T}$: list of binary question templates of all attributes \\
\hspace*{0.02in} {\bf Output:} $\operatorname{B}$: list of binary \textit{question-answer-image} triplets \\
\hspace*{0.02in} {\bf Initialization:} 
\begin{algorithmic}
\For{each specific attribute $a_k$}
\State $U_{a_k} \gets \{\}$: empty set of all image IDs with attribute $a_k$ 
\State $V_{a_k} \gets \{\}$: empty set of unique attribute values with attribute $a_k$
\EndFor
\State $C \gets \{\}$: empty set of unique categories 

\end{algorithmic}
\hspace*{0.02in} {\bf Build \textit{attribute-value-to-images} dictionary}: 
\begin{algorithmic}
\For{each fashion item $s_i \in \operatorname{S}$}
\State $C \gets c_i$
\For{each attribute $a_k \in s_i(\textrm{attributes})$}
\State $U_{a_k} \gets u_i$
\State $P_{c_i}$(image ID set of positive answer of category $c_i$) $\gets u_i$
\For{each attribute value $v_{a_k} \in s_i(\textrm{attribute values})$}
\State $V_{a_k} \gets v_{a_k}$; 
\State $P_{v_{a_k}}$(image ID set of positive answer with attribute value $v_{a_k}$) $\gets u_i $
\EndFor
\EndFor
\EndFor
\end{algorithmic}

\hspace*{0.02in} {\bf Build \textit{attribute-value-to-(positive/negative answer)-images} dictionary}: 
\begin{algorithmic}
\For{ each attribute value $v_{a_k} \in V_{a_k}$}
\State $V^\prime = \textrm{Synonyms}(v_{a_k})$ 
\For{ each $v_{+} \in (V^\prime \cap V_{a_k})$}
\State $P_{v_{a_k} }= P_{v_{a_k} }\cup P_{v_{+} }$
\EndFor
\EndFor
\For{ each attribute value $v_{a_k} \in V_{a_k}$}
\State $N_{v_{a_k}} = U_{a_k} - P_{v_{a_k}}$
\EndFor
\For{each category $c_i \in C$}
\State Follow the same strategy to update positive answer image ID set $P_{c_i}$ and build negative set $N_{c_i}$  
\EndFor
\end{algorithmic}

\hspace*{0.02in} {\bf Expand \textit{attribute-value-to-(positive/negative answer)-images} dictionary with attributes and category combinations}:
\begin{algorithmic}
\For{ each attribute value $v_{a_k} \in V_{a_k}$}:
\For{ each category $c_i \in C $}
\State $P_{(v_{a_k}, c_i)}$ : positive answer image ID set of the combination of $(v_{a_k}, c_i)$
\State $N_{(v_{a_k}, c_i)}$ : negative answer image ID set of the combination of $(v_{a_k}, c_i)$
\State $P_{(v_{a_k}, c_i)} = P_{v_{a_k}}\cap P_{c_i}$
\State $N_{(v_{a_k}, c_i)} = (P_{v_{a_k}}\cap N_{c_i}) \cup (N_{v_{a_k}}\cap N_{c_i} ) \cup (N_{v_{a_k}}\cap N_{c_i}$)
\EndFor
\EndFor
\end{algorithmic}
\hspace*{0.02in} {\bf Create balanced \textit{question-(positive/negative answer)-images} triplets}:
\begin{algorithmic}
\For{each category $c_i \in C$}
\For{each specific attribute $a_k$ of category $c_i$}
\State $\operatorname{Q_{T_{(a_k, c_i)}}} = \operatorname{Q_T}(\textrm{binary question templates of attribute} \, a_k \textrm{and category} \, c_i)$ 
\For{ each combination of attribute value $v_{a_k} \in V_{a_k}$ and category $c_i \in C$ }
\State $Q_{(v_{a_k}, c_i)} =$ Fill $\operatorname{Q_{T_{(a_k, c_i)}}}$ templates with $v_{a_k}$ and $c_i$ to generate binary questions
\For{ each binary question $q_{(v_{a_k}, c_i)} \in Q_{(v_{a_k}, c_i)}$}
\State Pick the same number of image IDs from $P_{(v_{a_k}, c_i)}$ and $N_{(v_{a_k}, c_i)}$:
\State $B \gets (q_{(v_{a_k}, c_i)}, \textrm{yes}, u_p \in P_{(v_{a_k}, c_i)}) \cup (q_{(v_{a_k}, c_i)}, \textrm{no}, u_n \in N_{(v_{a_k}, c_i)})$
\EndFor
\EndFor
\EndFor
\EndFor
\end{algorithmic}
\end{algorithm}

\textbf{Attribute-based:} First, we build an \textit{attribute-value-to-images} dictionary to map each distinct \textit{attribute value} or \textit{category} to a set of eligible image IDs. Given a specific \textit{attribute value}, we collect a set of positive answer image IDs directly from this \textit{attribute-value-to-images} dictionary using given \textit{attribute value} and its synonyms. The negative answer image IDs are collected from all image IDs of the same \textit{attribute} excluding the positive image IDs.
More concretely, to maximally reduce the noise in the positive/negative answer image IDs, we need to verify the relationship among \textit{attribute values} as alternative, hierarchical, or exclusive terms. Examples of alternative terminologies are ``sweatpants'', ``jogger pants'', and ``lounge pants''; examples of hierarchical terminologies are ``blue'', ``light blue'', and ``sky blue''; and, examples of exclusive terminologies are ``light blue'' and ``dark blue''. We expect \textit{attribute values} with similar terminologies (alternatives and parents of hierarchical terms) to contain the same set of positive samples, so they are considered synonyms. 
In this manner, we can build an \textit{attribute-value-to-(positive/negative answer)-images} dictionary (see Algorithm \ref{alg:balanced}).

Then, we consider all the combinations of assorted \textit{attributes} with \textit{category}. For example, \textit{$\langle$ color, pattern, category $\rangle$, $\langle$ color, category $\rangle$, $\langle$ material, neckline type, category $\rangle$}, etc. For each combination, we further expand the \textit{attribute-value-to-(positive/negative answer)-images} dictionary by mapping the combination of one specific \textit{attribute value} and one specific \textit{category} (e.g. $\langle$red, shirt$\rangle$) to its positive/negative answer image ID set. We collect the positive answer image ID set of the combinations following the formula in Equation \ref{eqn:pos} and the negative answer image ID set following the formula in Equation \ref{eqn:neg}:
\begin{align} 
  \label{eqn:pos}
  Pos(<attr1, category>) = & Pos(attr1)\cap Pos(category)\\
  \begin{split}
    \label{eqn:neg}
    Neg(<attr1, category>) = & (Pos(attr1)\cap Neg(category))\cup \\
    & (Neg(attr1)\cap Pos(category)) \cup \\
    & (Neg(attr1)\cap Neg(category))
  \end{split}
\end{align}
whereas, \textit{Pos()} is the positive answer image ID set and \textit{Neg()} is the negative answer Image ID set. With the \textit{attribute-value-to-(positive/negative answer)-images} dictionary, we can easily generate different binary questions via filling the question templates with each combination of \textit{attribute value} and \textit{category} in the dictionary. We can pick a fixed number of positive and negative answer image IDs to guarantee the sample balance for each question. Following the same formula, we can easily expand the combinations to multiple attribute values and one category. 

\begin{figure}
  \centering
  \includegraphics [width=\textwidth]{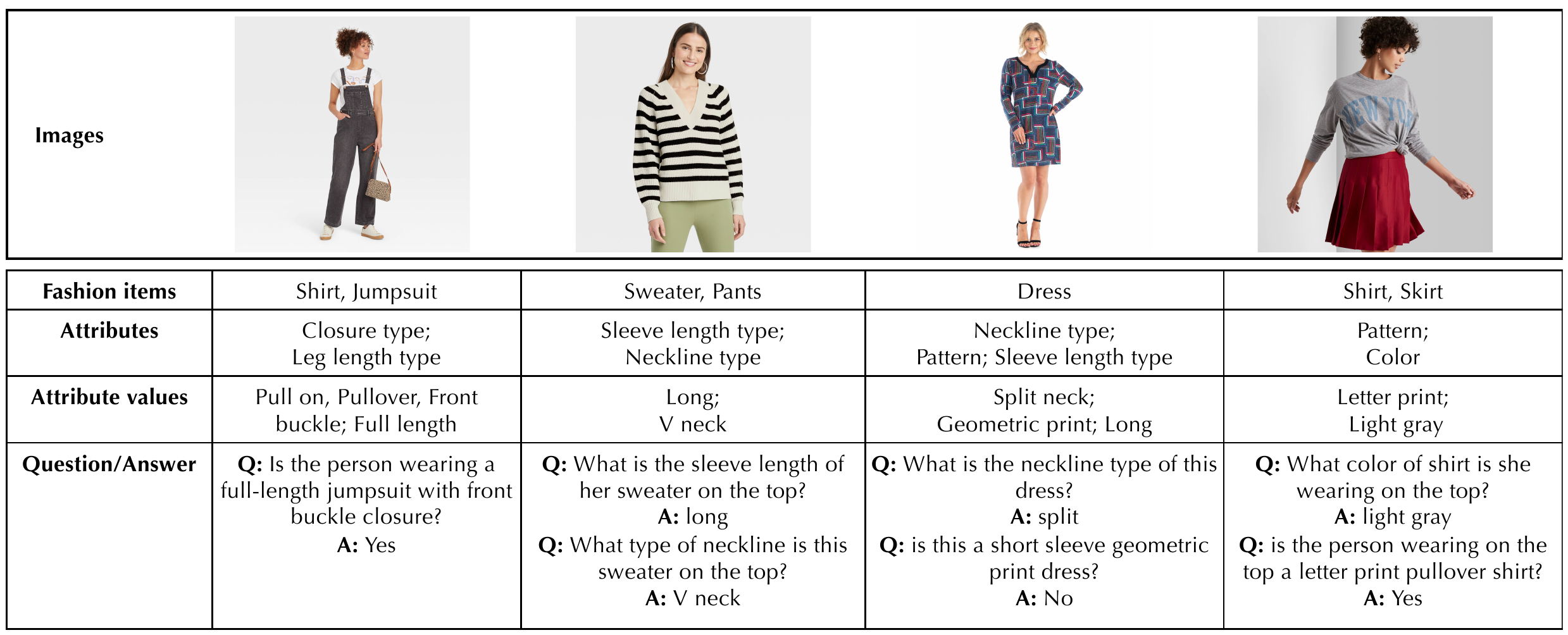}
  \caption{Four randomly picked \textit{question-answer-image} triplets from FashionVQA dataset.}
  \label{fig:dataset_snapshot}
\end{figure}
   
\subsection{Dataset description}
Figure~\ref{fig:dataset_snapshot} shows four randomly picked \textit{question-answer-image} triplet examples in our dataset. There are 42 \textit{attributes} in our dataset, including \textit{category, color, pattern, occasion, material, number of}, 29 type-related \textit{attributes}, 5 style-related \textit{attributes}, and 2 shape-related \textit{attributes}. The binary questions in our dataset are composed of three major types: \textit{category}, \textit{category + one attribute}, and \textit{category + two attributes} with 1, 2, and 6 permutations between \textit{category} and \textit{attribute}, respectively, along with the ascending difficulty level to learn the alignment between a given binary question and an input image. 

\textbf{FashionVQA:} FashionVQA dataset includes 207,654 unique photoshoot images with resolution $600 \times 600$. We use 169,406 images in the train split for training and 38,248 images in the validation split for evaluation. The train split is composed of 163M \textit{question-answer-image} triplets and the validation split includes 5.2M \textit{question-answer-image} triplets. Since the information in binary questions is much more complicated than that in the non-binary questions, there are more binary triplets than non-binary ones in the dataset. In the train split, we have 22M non-binary \textit{question-answer-image} triplets covering 33 different question types, and approximately 141M binary \textit{question-answer-image} triplets, among which 134M have questions with one \textit{category} and two \textit{attribute values}, 6M have questions with one \textit{category} and one \textit{attribute value}, and 1M have questions with only one \textit{category} or one \textit{attribute value}. In the validation split, we have 1.2M non-binary \textit{question-answer-image} triplets and 4M binary \textit{question-answer-image} triplets. The answer vocabulary contains 1,545 different classes in total.

\textbf{mini-FashionVQA:} We also create a subset dataset, named mini-FashionVQA, derived from the FashionVQA dataset. The mini-FashionVQA dataset includes 20M \textit{question-answer-image} triplets in the train split (11M from non-binary triplets and 9M from binary triplets) and 2.2M  triplets in the validation split (0.7M from non-binary triplets and 1.5M from binary triplets).
\section{Benchmarks}

\begin{figure}
  \centering
  \includegraphics [width=\textwidth]{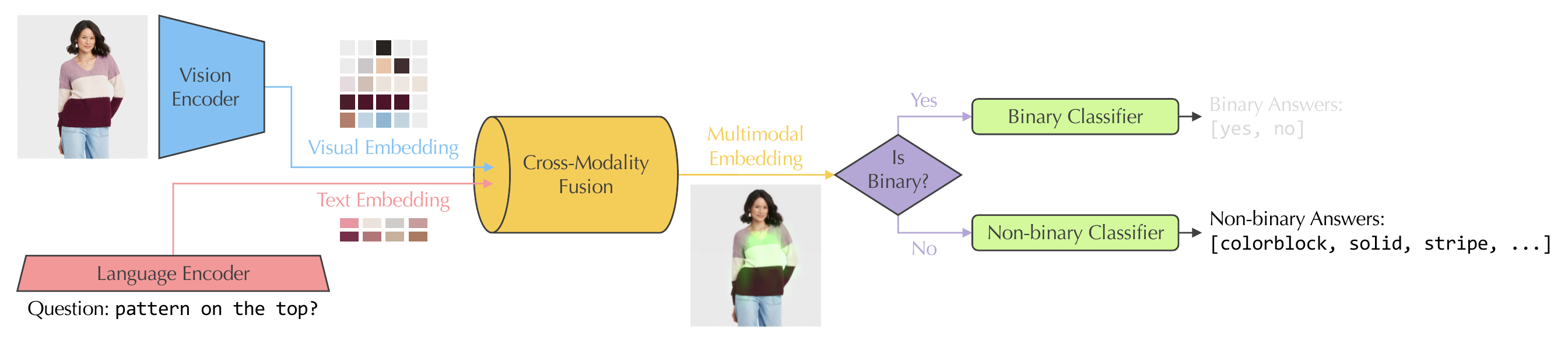}
  \caption{Pipeline of the fashion VQA task.}
  \label{fig:pipeline}
\end{figure}

Every benchmark reported on our datasets is implemented via PyTorch\cite{pytorch}-v1.10 on servers with 8 Nvidia 80GB A-100 GPUs, 2 AMD 2.25GHz 7742 CPUs, and 4TB system memory. In the training stage, we adopted data-parallel multi-GPU training and set the batch size to 2048, and trained for 40 epochs. The Adam\cite{adam} optimizer is used across all the models. The learning rate is set to 0.0001 and reduced by half at the milestone epochs of 20, 30, and 35.

We benchmark the FashionVQA dataset by training several VQA models to learn the interaction between images and questions. Figure~\ref{fig:pipeline} shows the VQA pipeline adopted in our experiment. Given the visual embedding of the input image and text embedding of the input question sentence, we train the model to output the given answer to the question. The dataset is used to train two variants of the MCAN \cite{mcan} model and a MUTAN \cite{mutan} model. One MCAN variant, named MCAN$\sp{\ast}$-v1, is a modification of the MCAN-small, which includes only two encoder-decoder modules. The other variant is named MCAN$\sp{\ast}$-VLM, which has a similar structure to MCAN$\sp{\ast}$-v1, but instead of an answer classifier, it has a token classifier covering all of the question and answer tokens. For MCAN$\sp{\ast}$-VLM, the answer to each question is tokenized as one token and concatenated with the question tokens as the language input. The special token `SEP' is inserted between the question and the answer. Also, `EOS' token is used at the end of the answer. During the training of MCAN$\sp{\ast}$-VLM, we randomly mask one token and predict the masked token as in the masked language modeling, similar to BERT \cite{bert}. Different from MCAN$\sp{\ast}$-v1 that answer vocabulary is independent of the word vocabulary of the questions, MCAN$\sp{\ast}$-VLM maps each answer to one token and expands the original word vocabulary to a larger one with the answer tokens. Thus, the tokens in the answers and questions share the same word vocabulary. This allows the MCAN$\sp{\ast}$-VLM to work as a visual language model, which directly benefits from the overlap in the question tokens of the binary questions and the answer tokens of the non-binary question.

In the training stage, except MCAN$\sp{\ast}$-VLM, we treat the binary-question prediction and the non-binary question prediction as two different tasks and output the predicted answers from two different classifiers. We report top-1 accuracies for both binary and non-binary samples.

 \begin{table}[h]
  \caption{Benchmarks of MCAN$\sp{\ast}$-v1, MCAN$\sp{\ast}$-VLM, and MUTAN trained on FashionVQA dataset}
  \label{full}
  \begin{center}
  \begin{tabular}{c|ccc}
      \hline
      \multirow{2}{*}{Model} &\multicolumn{3}{|c}{Top-1 Acc}  \\ \cline{2-4}
      &  All & Non-binary & Binary \\ \hline
      MUTAN & 81.38\%&61.62\%&87.43\% \\
      MCAN$\sp{\ast}$-v1 & 84.42\%&64.32\%&90.58\% \\ 
      MCAN$\sp{\ast}$-VLM & \bf{84.69\%}&
\bf{64.65\%}&
\bf{90.84\%}
\\ \hline 
  \end{tabular}
\end{center}
\end{table}

Table~\ref{full} lists the benchmark results of the three aforementioned models on the validation split of our FashionVQA dataset. The results show that MCAN$\sp{\ast}$-VLM works better than MCAN$\sp{\ast}$-v1 and MUTAN, indicating that a decoder-only visual language model (VLM) performs better than the dedicated VQA architectures.

\begin{figure}
  \centering
  \includegraphics [width=\textwidth]{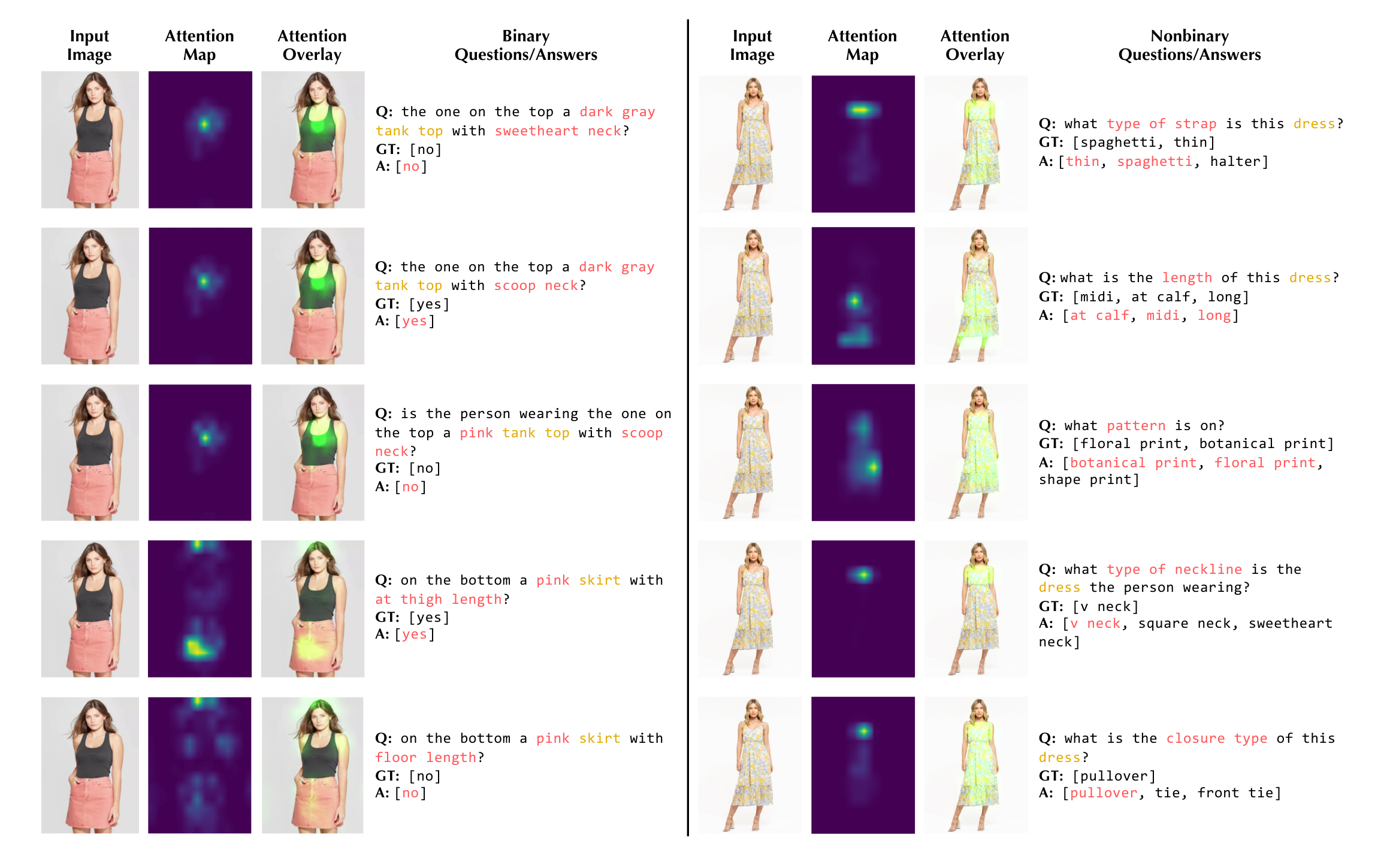}
  \caption{Visualization of attention maps generated by the model trained with FashionVQA dataset.}
  \label{fig:att_v}
\end{figure}

By visualizing the image attention maps generated from an intermediate layer of the model, we can validate whether the model focuses its attention on the regions mentioned in a question. Figure~\ref{fig:att_v} visualizes the attention map from two validation samples for a series of binary and non-binary questions. The three columns of images on each side are the input images, attention maps, and images overlayed by the attention maps, respectively, followed by the corresponding input questions, ground truth answers, and predicted answers. When the questions focus on different fashion items of the same image, the attention map shifts to the targeted region as expected.

\subsection{Benchmarks with different VQA models}
We also use the mini-FashionVQA to benchmark a larger variety of VQA models including Bottom-up-top-down (BUTD) \cite{bottom-up-attention}, MUTAN \cite{mutan}, DFAF \cite{dfaf}, MCAN$\sp{\ast}$-v1, MCAN$\sp{\ast}$-v2, MCAN$\sp{\ast}$-VLM, and OSCAR \cite{oscar}. MCAN$\sp{\ast}$-v2 has the same model structure as MCAN$\sp{\ast}$-v1 except for its intermediate feed-forward layer with only half the number of channels of MCAN$\sp{\ast}$-v1. We apply similar visual embedding, text embedding, and loss function in these models and train them from scratch. 

Table~\ref{comp-models-reg} lists the results (average of top-1 accuracies for three runs) from different VQA models with the same region-based visual features as input. The visual features are extracted from Faster-RCNN with ResNet-101 backbone fine-tuned with VisualGenome. We set the maximum number of objects extracted from the object-detection model to 25. The feature dimension of each object is 2048. A combination of GLove \cite{glove} + GRU\cite{GRU}/LSTM\cite{LSTM} is used for the text embedding in DFAF, MCAN$\sp{\ast}$-v1, MCAN$\sp{\ast}$-v2, MCAN$\sp{\ast}$-VLM, and BUTD. MUTAN adopts GRU for the text embedding, of which the parameters are initialized with SkipThoughts \cite{skipthoughts}. The number of parameters, FLOPs, and activation counts in all our experiments are calculated only from the cross-modality models, excluding text embedding and visual embedding components. On the mini-FashionVQA dataset, MCAN$\sp{\ast}$-VLM achieves the best accuracy for both non-binary question and binary question samples, with fewer parameters and FLOPs than OSCAR. Also, MCAN$\sp{\ast}$-VLM works better than MCAN$\sp{\ast}$-v1 on both non-binary questions and binary questions.      

\begin{table}[ht]
  \caption{Performance on different VQA models trained on the mini-FashionVQA dataset with same region-based visual features}
  \label{comp-models-reg}
  \begin{center}
  \begin{tabular}{ccccccc}
      \hline
      \multirow{2}{*}{Model}&\multirow{2}{*}{Parameters}&
      \multirow{2}{*}{FLOPs}& \multirow{2}{*}{Act.Count}&
      \multicolumn{3}{c}{Top-1 Acc} \\ \cline{5-7}
      & & & & All & Non-binary& Binary \\ \hline
      MUTAN &9.8M &38.5M &156.9K& 75.08\%&59.14\%&81.50\%
   \\  
      BUTD &11.5M &61.5M &30.7K&79.26\%&63.61\%&85.56\%
 \\ 
      DFAF & 9M& 280M& 114.8K&80.55\%&62.52\%&87.81\%
 \\ 
   OSCAR &86.7M &6475M &2832.3K &81.21\%&
64.20\%&
88.05\%

 \\ \hline 

      MCAN$\sp{\ast}$-v1 & 19M &427M &238.9K & 81.69\% & 64.47\% & 88.61\%
\\  
      MCAN$\sp{\ast}$-v2 & 14.5M &320M &134.5K &81.08\%&64.33\%&87.83\%
 \\ 
       MCAN$\sp{\ast}$-VLM & 19M &464M &282.0K &\bf{81.80\%}&\bf{64.63\%}&\bf{88.71\%}
  \\ \hline
      \end{tabular}
\end{center}
\end{table}

\subsection{Ablation study}
\textbf{Impact of visual embedding extraction schemes:} We also benchmark different visual embedding extraction schemes and see their impact on the performance of VQA tasks for the mini-FashionVQA dataset. We replace the region-based feature with the grid feature with the same dimension. The grid-feature (refer to \cite{grid-feature}) model is built with ResNext-101 backbone and fine-tuned with object detection task on the VisualGenome dataset. The visual feature is extracted from the same layer as the object detection and pooled into different sizes. To be aligned with the visual input dimension size from the region-based feature, the spatial dimension of the grid feature is set to 5\-$\times$\-5 with the feature dimension set to 2048. Other than the visual embedding, all of the settings remain the same.

\begin{table}[ht]
 \caption{Performance with region-based and grid visual features across different VQA models}
  \label{re-grid}
  \begin{center}
  \scalebox{0.9}{
  \begin{tabular}{c|ccc|ccc}
      \hline
      \multirow{2}{*}{Model}&\multicolumn{3}{|c|}{Region-based features (ResNet-101)} &
      \multicolumn{3}{c}{Grid features (ResNext-101)} \\ \cline{2-7}
      & All&Non-binary &Binary & All & Non-binary& Binary \\ \hline
      MUTAN  &75.08\%&59.14\%&81.50\%& 79.77\%\small{(+4.69\%)}&62.54\%&86.70\% \\  
      BUTD  & 79.26\%&63.61\%&85.56\%&80.54\%\small{(+1.28\%)}&64.30\%&87.08\% \\
      DFAF  &80.55\%&62.52\%&87.81\%&82.01\%\small{(+1.46\%)}&64.70\%&88.97\% \\ \hline
      MCAN$\sp{\ast}$-v1 & 81.69\%& 64.47\%&88.61\% & 
        83.29\%\small{(+1.60\%)}&65.38\%&90.49\%
       \\ 
  MCAN$\sp{\ast}$-v2 & 81.08\%&64.33\%&87.83\%& 82.98\%\small{(+1.90\%)}&65.17\%&90.14\% \\  
 MCAN$\sp{\ast}$-VLM & 81.80\%&64.63\%&88.71\%&
 83.41\%\small{(+1.61\%)}&
65.52\%&
90.62\%

 \\  \hline
  \end{tabular}
} 
\end{center}
\end{table}

Table~\ref{re-grid} shows that the grid-feature-based visual embedding extraction method consistently works better than the region-based method across all different VQA models by more than 1\% when trained on our dataset. In the other experiments, unless mentioned otherwise, we use grid-feature-based visual embedding for all the models.

\textbf{Impact of different backbones for visual embedding:} Generally, a better visual backbone will contribute to better visual embedding. We benchmark three different visual backbones (ResNet-50, ResNext-101, ResNext-152) for the grid-feature extraction on our dataset for MCAN$\sp{\ast}$-v1, BUTD, and DFAF. All the visual backbones are pre-trained on VisualGenome \cite{visualgenome} dataset for the grid-feature extraction.

\begin{table}[ht]
\caption{Performances with different visual backbones for grid-feature}
  \label{comp-backbones}
  \begin{center}
  \scalebox{0.8}{
  \begin{tabular}{c|ccc|ccc|ccc}
      \hline
      \multirow{2}{*}{Model}&\multicolumn{3}{|c|}{MCAN$\sp{\ast}$-v1} &\multicolumn{3}{|c|}{BUTD} &\multicolumn{3}{|c}{DFAF}\\ \cline{2-10}
       & All&Non-binary &Binary & All & Non-binary& Binary & All & Non-binary& Binary \\ \hline
       ResNet-50&82.99\%&
65.25\%&
90.13\%&
80.47\%&
64.35\%&
86.96\%&
80.81\%&
63.39\%&
87.81\%
\\ 
ResNext-101 &
83.29\%&
65.38\%&
90.49\%&
80.54\%&
64.30\%&
87.08\%&
82.01\%&
64.70\%&
88.97\%  \\ 
ResNext-152 & 83.15\%&
65.41\%&
90.29\%&
80.50\%&
64.62\%&
86.89\%&
82.39\%&
65.17\%&
89.32\% \\ \hline
  \end{tabular}}
\end{center}
\end{table}

Table~\ref{comp-backbones} shows that ResNext-101 constantly works better than ResNet-50 on three different models for the performance of both non-binary and binary questions; however, the performance improvement from ResNext-101 to ResNext-152 is inconsistent. Overall, grid-feature with ResNext-101 as the backbone is the best choice for extracting visual features on our dataset.

\textbf{Impact of different spatial dimension sizes for grid feature:} A larger spatial dimension size after the pooling operation will typically preserve more visual information. We benchmark MCAN$\sp{\ast}$-v1 with three different spatial dimension sizes (5\-$\times$\-5, 7\-$\times$\-7, and 9\-$\times$\-9) for the grid feature visual embeddings in Table~\ref{mcan-sp}. ResNext-101 is the selected visual backbone.

The results in Table~\ref{mcan-sp} show that the best performance among the three is from the smallest spatial dimension size, 5\-$\times$\-5, rather than the largest one. One possible reason is that the background of the photoshoot images from our dataset includes some trivial information, and the larger spatial dimension sizes do not add useful information.

  \begin{table}[ht]
  \caption{Performances with different spatial dimension sizes for grid-feature}
  \label{mcan-sp}
  \begin{center}
  \begin{tabular}{ccccc}
      \hline
      \multirow{2}{*}{Model} &\multirow{2}{*}{Spatial dimension size}&\multicolumn{3}{c}{Top-1 Acc}  \\ \cline{3-5}
      &  & All & Non-binary & Binary \\ \hline
      \multirow{3}{*}{MCAN$\sp{\ast}$-v1} & 5\-$\times$\-5 &
      83.29\%&65.38\%&90.49\% \\
     
      & 7\-$\times$\-7 &82.94\%&64.19\%&90.48\%\\
      & 9\-$\times$\-9 &82.60\%&65.23\%&89.59\%
 \\ \hline 
  \end{tabular}
\end{center}
\end{table}

\textbf{Impact of single-task versus multi-task training:} Due to the large difference in the answer distribution of non-binary questions and binary questions, we consider using different classifiers for answer predictions and treating the problem as a multi-task classification. Namely, predicting answers for two types of questions with either a single classifier or two separate classifiers. This applies to all models, except the MCAN$\sp{\ast}$-VLM model, where the outputs are generated by a single token classifier, including both answer and question tokens.

\begin{table}[ht]
\caption{Performance with different number of classifiers for non-binary and binary questions}
  \label{comp-classifiers}
  \begin{center}
  \begin{tabular}{c|ccc|ccc}
      \hline
      \multirow{2}{*}{Model}&\multicolumn{3}{|c|}{Single-task} &
      \multicolumn{3}{|c}{Multi-tasks} \\ \cline{2-7}
      & All&Non-binary &Binary & All & Non-binary& Binary \\ \hline
       MUTAN  & 79.40\%&
62.12\%&
86.36\%&
79.77\%\small(+0.37\%)&
62.54\%&
86.70\%\\ 
     BUTD & 80.32\%&
63.55\%&
87.07\%&
80.54\%\small{(+0.22\%)}&
64.30\%&
87.08\% \\ 
      DFAF &81.6\%&
63.85\%&
88.74\%&
82.01\%\small{(+0.41\%)}&
64.70\%&
88.97\% \\ 
MCAN$\sp{\ast}$-v1 & 
83.24\% &
65.36\% &
90.44\%&
83.29\%\small{(+0.05\%)}&65.38\%&90.49\%
 \\ \hline 
       \end{tabular}
  
\end{center}
\end{table}

Table~\ref{comp-classifiers} demonstrates that the proposed multi-task classification is superior to a single-task classification in predicting the answers for the VQA models.

\section{Comparison to human performance}

\textbf{Human accuracy for FashionVQA dataset:} To see how well humans can answer the question in our dataset, we implemented a user interface that shows one \textit{question-image} pair from the validation set at a time. The user interface allows the human annotators to select one of the acceptable answers among 1,545 answer classes, e.g., ``yes'', ``no'', ``purple'', ``unicorn print'', ``tailored'', ``fly hook and loop fastener'', ``three quarter length'', etc. We asked the annotators to answer each question to the best of their knowledge without looking up the terms.

We have two types of annotators: experts and non-experts. We trained our expert annotators with at least ten examples per fashion term in our word vocabulary. Both expert and non-expert annotators are trained on the VQA task of our dataset. Table~\ref{human-on-dataset} shows the accuracies of nine human annotators compared to the MCAN$\sp{\ast}$-VLM model trained on the FashionVQA dataset.

\begin{table}[ht]
  \caption{Performances of different human annotators on samples from FashionVQA validation set}
  \label{human-on-dataset}
  \begin{center}
    \scalebox{0.8}{
      \begin{tabular}{l|ccc|ccc|ccc}
        \toprule
         \multirow{2}{*}{Annotator}&\multicolumn{3}{|c|}{Number of samples}&\multicolumn{3}{|c|}{Accuracy} &\multicolumn{3}{c}{Accuracy \textit{p}-value} \\  \cline{2-10}

         &   All &   Non-binary  &   Binary &   All &   Non-binary &   Binary &   All &   Non-binary&   Binary \\ \hline
        
 Expert 1          &       728 &                  216 &              512 &      63.6\% &                 43.5\% &             72.1\% &            8.5e-30 &                       1.1e-09 &                   7.5e-20 \\
 Non-expert 1          &       106 &                   29 &               77 &      58.5\% &                 24.1\% &             71.4\% &            1.8e-07 &                       1.4e-05 &                   0.00018 \\
 Non-expert 2        &        70 &                   18 &               52 &      52.9\% &                 22.2\% &             63.5\% &            6.9e-07 &                        0.0003 &                   8.7e-05 \\
 Non-expert 3          &        61 &                   17 &               44 &      63.9\% &                 29.4\% &             77.3\% &            0.00072 &                        0.0035 &                      0.02 \\
 Non-expert 4         &        51 &                   14 &               37 &      47.1\% &                 14.3\% &             59.5\% &            1.2e-06 &                       8.7e-05 &                   0.00025 \\
 Non-expert 5          &       150 &                   44 &              106 &      50.7\% &                 22.7\% &             62.3\% &              3e-14 &                       2.8e-08 &                   1.3e-08 \\
 Non-expert 6        &       211 &                   62 &              149 &      52.6\% &                 22.6\% &             65.1\% &            9.5e-18 &                       3.9e-11 &                   4.4e-10 \\
 Non-expert 7        &       103 &                   27 &               76 &      48.5\% &                 25.9\% &             56.6\% &            3.3e-11 &                       6.2e-05 &                   3.6e-08 \\
 Non-expert 8         &        50 &                   14 &               36 &      52.0\% &                 14.3\% &             66.7\% &            1.6e-05 &                       8.7e-05 &                    0.0023 \\
        \bottomrule
      \end{tabular}
    }
  \end{center}
\end{table}

To analyze the statistical significance of the results, we calculated the \textit{p}-values of the human accuracies with respect to the validation accuracy of the model using the one-sided t-test. The validation accuracies of the MCAN$\sp{\ast}$-VLM model are 84.69\%, 64.65\%, and 90.84\% for all, non-binary, and binary questions, respectively. The model outperforms all of the human annotators, and at a 95\% confidence level, the differences between the model validation accuracy and human accuracies are statistically significant.

\textbf{Accuracies for human-generated questions:} We also stress-tested the model by measuring its performance on human-generated questions. We asked an expert annotator, Expert 2, to paraphrase the questions of 300 random samples (218 binary and 82 non-binary samples) from the validation set. We used these questions instead of the original questions in the validation set to measure the accuracies of the MCAN$\sp{\ast}$-VLM model and a human annotator, Expert 1, as shown in Table~\ref{paraphrased}.
\begin{figure}[htp]
  \centering
  \includegraphics [width=0.95\textwidth]{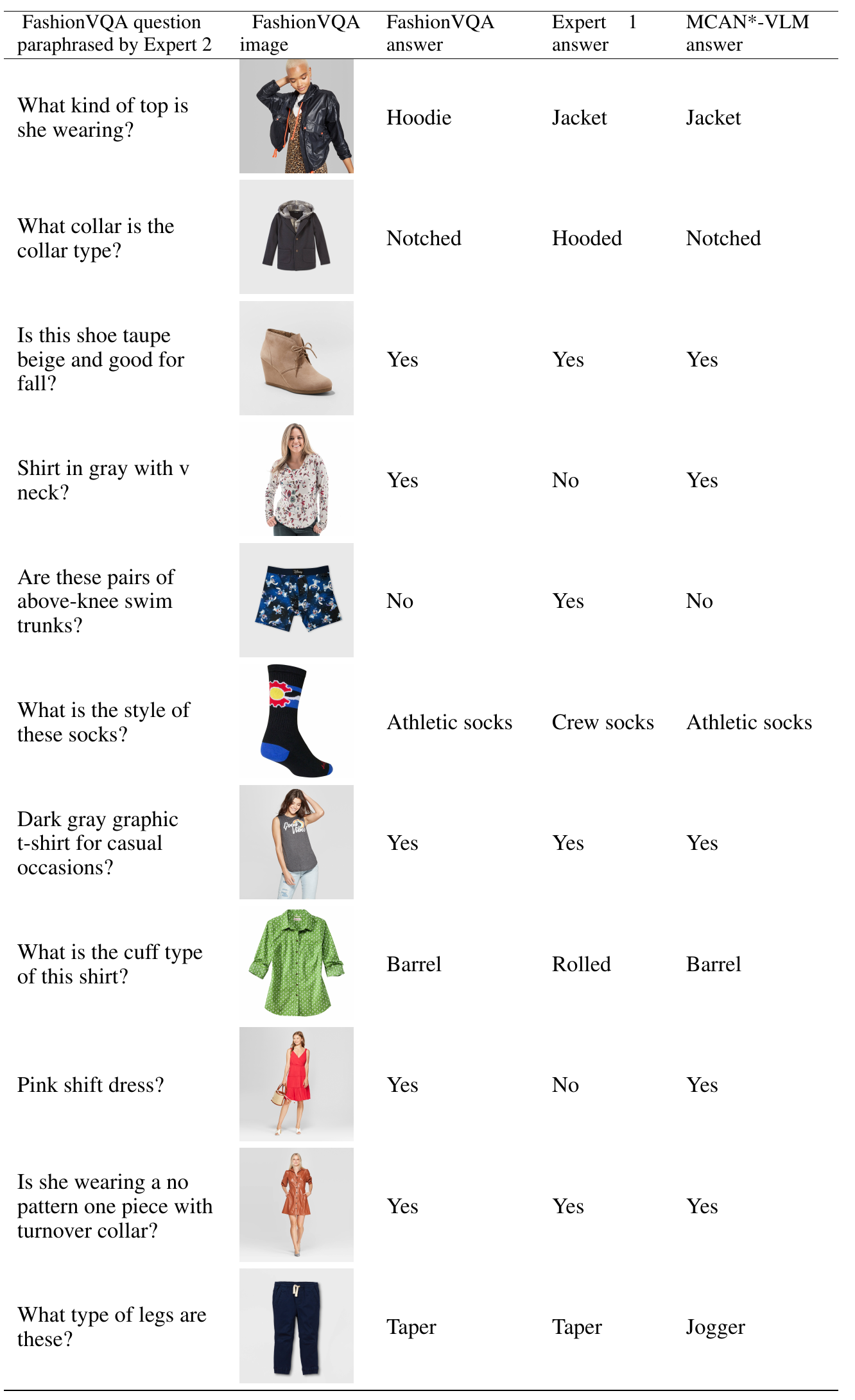}
  \caption{FashionVQA paraphrased and answered by humans and model}
  \label{fig:paraphrased-examples}
\end{figure}

\begin{table}[ht]
  \caption{Performances of the MCAN$\sp{\ast}$-VLM model and a human expert on human-generated questions}
  \label{paraphrased}
  \begin{center}
    \scalebox{1.0}{
      \begin{tabular}{lrrrrrrrrr}
        \toprule
          \multirow{2}{*}{}&\multicolumn{3}{c}{Accuracy} \\ \cline{2-4}
       & All & Non-binary & Binary \\ \midrule
        Human Expert 1 &   62.3\% &   30.5\% &    74.3\%  \\
        MCAN$\sp{\ast}$-VLM &   77.7\% &   47.6\% &    89.0\%  \\
        \midrule
        \textit{p}-value &    1.9e-05 &  0.0125  &  3.4e-05  \\
        \bottomrule
      \end{tabular}
    }
  \end{center}
\end{table}

\begin{figure}
  \centering
  \includegraphics [width=\textwidth]{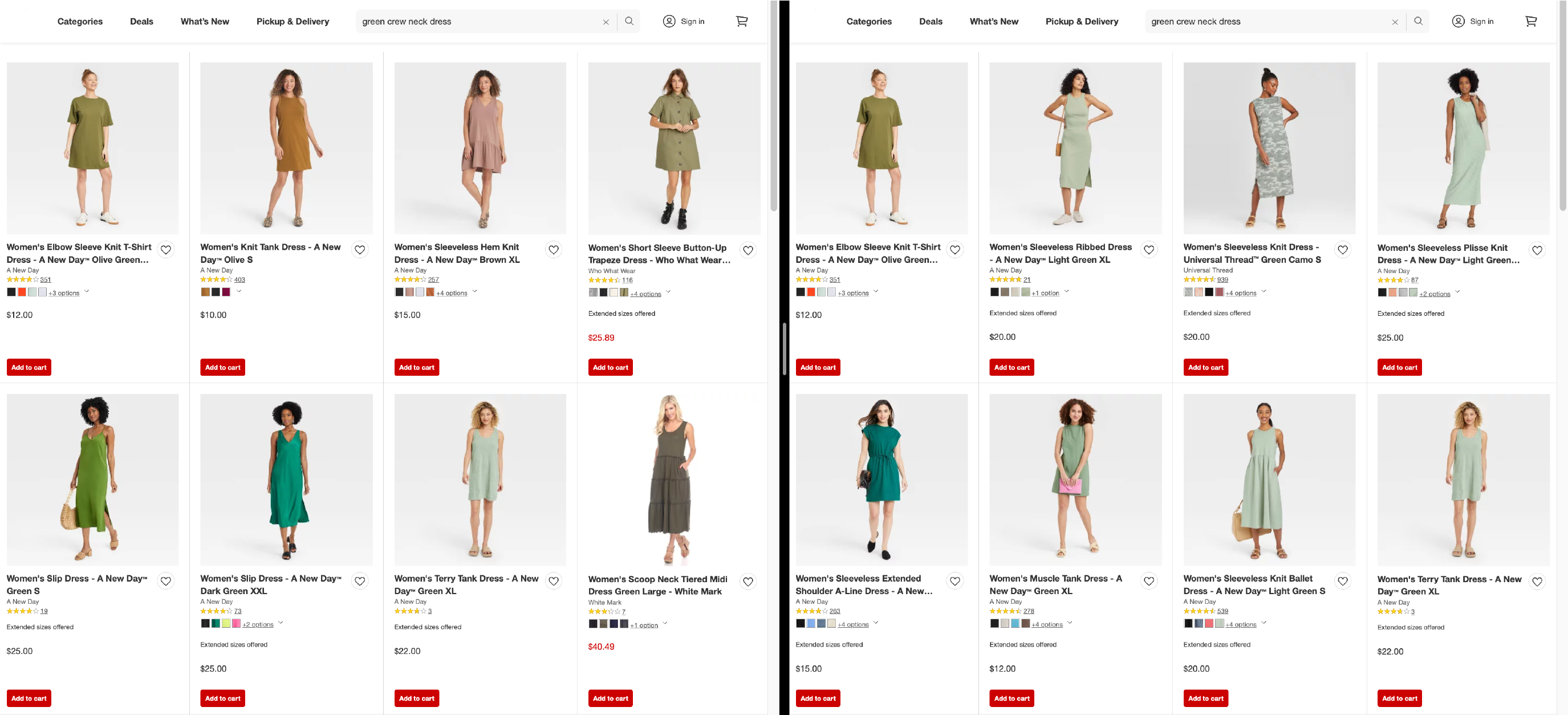}
  \caption{An example of the side-by-side comparison of the search results with and without reranking. Given a random search query, e.g. ``green crew neck dress'', the annotator picks her/his preferred search results between the left (A) and right (B) result pages.}
  \label{fig:ab_test}
\end{figure}

We performed a one-sided t-test to analyze the statistical significance of the difference between the human and the model accuracies. At a significance level of 0.05 ($\alpha=0.05$), the \textit{p}-values reject the null hypothesis of the human accuracy being greater than or equal to the model. Figure~\ref{fig:paraphrased-examples} provides several examples from this experiment.
 
\textbf{Impact on downstream tasks:} We performed a side-by-side comparison of the apparel search with and without FashionVQA. A baseline search engine returns the top 24 items for an apparel search query. Another variant of the search results is formed by reranking these 24 items with FashionVQA: we generate a set of binary questions from the search query and use MCAN$\sp{\ast}$-VLM model trained with FashionVQA to answer these questions for each of the 24 items. The average confidence scores of the yes and no answers are used as the additional features to rerank the top 24 items.

For a number of randomly-selected search queries with two \textit{attribute values} and one \textit{category}, e.g., ``green crew neck dress'', a human annotator is presented with the original and reranked search result pages and gets to choose her/his preferred result page. The result pages are randomly located on the left and right sides of the screen without the annotator knowing which of the two pages presents the reranked results. Figure~\ref{fig:ab_test} shows an example of our side-by-side A/B test for the given random query. Out of 150 search queries, the human annotator preferred 117 search pages reranked based on the FashionVQA. Binomial statistical test results in a \textit{p}-value of 3.2e-12, showing that the human annotator significantly prefers the search result page reranked using FashionVQA.

\section*{Conclusion}
In this work, we design a fashion VQA dataset and generate non-binary and binary questions via diverse templates. The templates allow us to flexibly scale the dataset to the size and complexity required for training a domain-specific multimodal model. We benchmark this large-scale dataset on different VQA models and discuss several factors impacting the performance of the VQA task. The best model is a visual language model trained on the FashionVQA dataset. The model generates the cross-modality embeddings of the vision and language domains applicable to downstream tasks of fashion dialogue, search, and recommendation.

\bibliographystyle{unsrtnat}
\bibliography{refs}

\end{document}